\documentclass{article}

 \usepackage[dblblindworkshop, final]{neurips_2025}
\workshoptitle{Efficient Reasoning}

\usepackage[utf8]{inputenc} 
\usepackage[T1]{fontenc}    
\usepackage{url}            
\usepackage{booktabs}       
\usepackage{amsfonts}       
\usepackage{nicefrac}       
\usepackage{microtype}      
\usepackage{xcolor}         

\usepackage{graphicx}
\usepackage{wrapfig}
\usepackage{fancyvrb}
\usepackage{fvextra}

\usepackage{amsmath}
\usepackage{amssymb}
\usepackage{mathtools}
\usepackage{amsthm}

\usepackage[capitalize,noabbrev]{cleveref}
\usepackage{microtype}
\usepackage{graphicx}
\usepackage{subfigure}
\usepackage{booktabs} 

\usepackage{graphicx}
\usepackage{xcolor}
\usepackage{listings}
\usepackage{enumitem}
\usepackage{caption}
\usepackage{multirow}
\usepackage{subfigure}
\usepackage{threeparttable}
\usepackage{wrapfig}

\usepackage{algorithm}
\usepackage{algpseudocode}
\usepackage{enumitem}

\workshoptitle{Efficient Reasoning}

\title{SpatialTraceGen: High-Fidelity Traces for Efficient VLM Spatial Reasoning Distillation}

\author{%
  Gio Huh\thanks{Equal contribution (order determined via dice roll)}\hspace{0.3em}, Dhruv Sheth\textsuperscript{*}, Rayhan Zirvi\textsuperscript{*}, Frank Xiao\textsuperscript{*} \\
  Computing + Mathematical Sciences\\
  California Institute of Technology\\
  \texttt{\{ghuh, dsheth, rayhanzirvi, frank\}@caltech.edu} \\
}

\begin{document}

\maketitle

\begin{abstract}
  While Vision-Language Models (VLMs) excel in many areas, they struggle with complex spatial reasoning, which requires problem decomposition and strategic tool use. Fine-tuning smaller, more deployable models offers an efficient path to strong performance, but this is hampered by a major bottleneck: the absence of high-quality, step-by-step reasoning data. To address this data-efficiency gap, we introduce SpatialTraceGen, a framework to distill the reasoning processes of a large teacher model into a high-quality dataset of multi-hop, multi-tool reasoning traces. A key innovation is our automated Verifier, which scalably ensures the fidelity of each reasoning step, providing a cost-effective alternative to manual human annotation. On the CLEVR-Humans benchmark, this verifier-guided process improves the average quality score of traces by 17\% while reducing quality variance by over 40\%. SpatialTraceGen delivers a dataset of expert traces, providing the structured, step-by-step examples of tool use necessary for effective fine-tuning and sample-efficient offline reinforcement learning. The code and dataset can be found at \url{https://anonymous.4open.science/r/spatial_trace-8535/}
\end{abstract}

\section{Introduction}

\begin{figure}[ht!]  
  \centering
  \includegraphics[width=\linewidth]{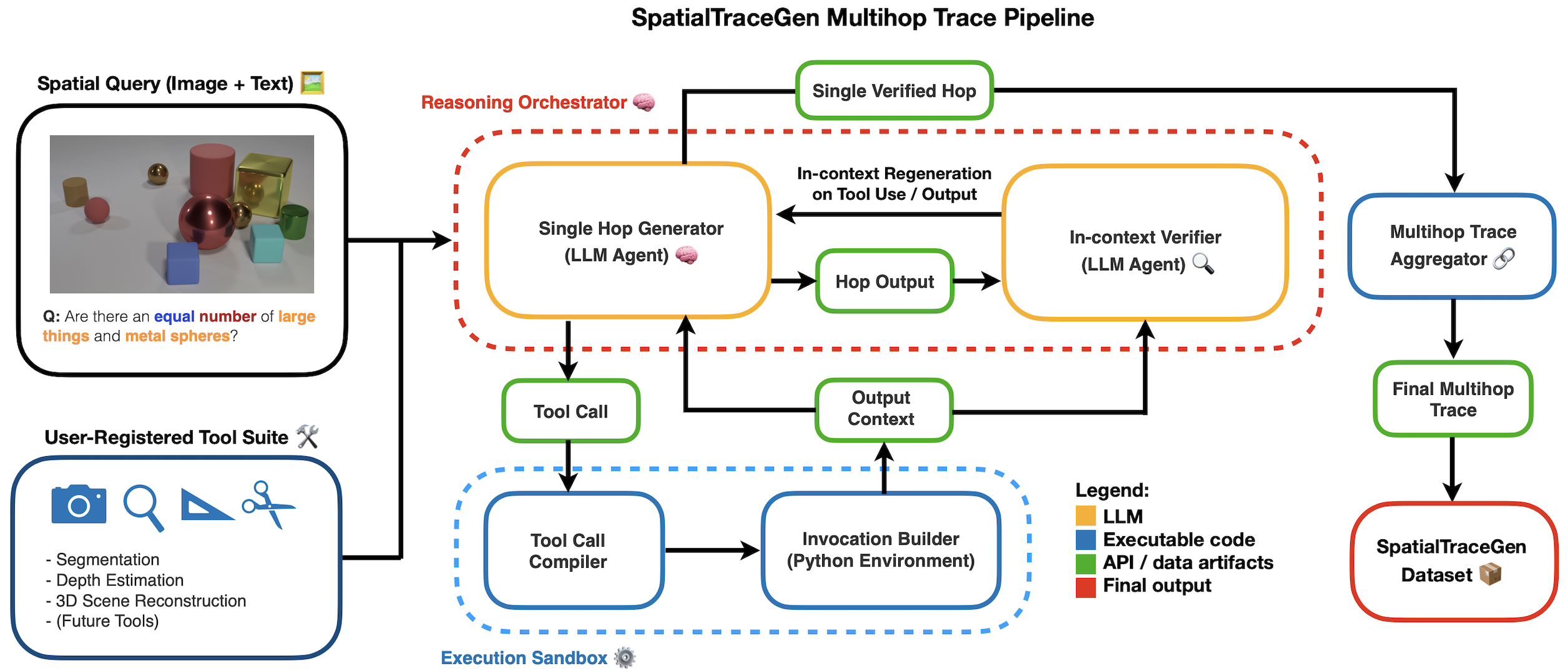}
  \vspace{-0.5em}
    \caption{\textbf{SpatialTraceGen pipeline.} The Single Hop Generator (yellow) breaks spatial queries into steps, invokes vision tools (blue), and records traces. A Verifier LLM validates each step before inclusion in the SpatialTrace corpus (red). Green boxes show API/data flow.}  \label{fig:spatialtrace-pipeline}
\end{figure}

Spatial reasoning, the ability to reason about objects and their interactions in space, remains a significant challenge for state-of-the-art Vision-Language Models (VLMs). Leading models perform near random chance on various spatial tasks \citep{stogiannidis2025gap}, with even GPT-4 showing critical failures in spatial cognition \citep{yang2025spatialcognition}. These shortcomings stem from architectural imbalances where models overweight textual priors over visual grounding \citep{chen2025spatialreasoninghardvlms}. While augmenting VLMs with external tools shows promise, progress is stalled by a critical data bottleneck: existing datasets lack high-fidelity, multi-hop demonstrations of problem decomposition, strategic tool selection, and information synthesis required for complex spatial reasoning.

Existing approaches have significant limitations for generating quality training data. Data-centric solutions like SpaRE \citep{ogezi2025spare} and Perspective-Aware Reasoning \citep{lee2025perspectiveaware} focus on static pattern recognition rather than dynamic, multi-step problem-solving. Agentic frameworks like OctoTools \citep{lu2025octotools} are designed for direct task execution, not pedagogical data generation, while others like VADAR \citep{marsili2025visualagentic} struggle with extensive multi-step reasoning. RL-based approaches like ReTool \citep{feng2025retool} and SWiRL \citep{goldie2025synthetic} are often constrained to single tools and produce noisy traces that degrade supervised fine-tuning performance or create "learning traps" \citep{Su2025PixelReasoner}.

This data scarcity creates an efficiency gap, preventing fine-tuning of smaller, deployable models on the complex reasoning patterns of larger systems. To address this, we introduce \textbf{SpatialTraceGen}, a framework for distilling complex reasoning capabilities into high-quality expert trace datasets. Our system orchestrates a VLM agent with diverse vision tools and introduces an automated Verifier for scalable quality assessment, replacing expensive human annotation. The resulting traces are structured as state-action-reward tuples, enabling direct compatibility with sample-efficient offline reinforcement learning.

Our primary contributions are:
\begin{itemize}
   \setlength{\itemsep}{1pt}
   \item \textbf{Data Generation Framework.} A system orchestrating LLM agents with vision tools to generate verifier-vetted reasoning chains for scalable, high-fidelity training data creation.
   \item \textbf{Knowledge Distillation Dataset.} Complex reasoning traces designed to transfer spatial reasoning capabilities from large to smaller models through fine-tuning.
   \item \textbf{Verifier-Driven Quality Improvement.} Empirical validation showing our automated verifier improves average trace quality by 17\% and reduces variance by over 40\% without sacrificing accuracy.
\end{itemize}

\section{Approach}
\label{sec:approach}

To enable efficient knowledge distillation from large, proprietary models to smaller, open-source ones, we shift the paradigm from direct task execution to data generation. Our framework, \textbf{SpatialTraceGen}, generates high-quality datasets of expert reasoning traces by formalizing spatial reasoning as a sequential decision-making process, making the data immediately compatible with modern efficient fine-tuning techniques.

\subsection{Reasoning Traces as MDP formulation}
\label{subsec:rl_environment}
We formulate the generated data as an offline RL environment, providing a structured basis for training policies without costly online interaction. A generated reasoning trace $\gamma$ is a sequence of state-action-reward tuples:
$$ \gamma = (s_0, a_0, r_0, s_1, a_1, r_1, \dots, s_T, a_T, r_T) $$
This formulation enables the direct application of offline RL algorithms (e.g., IQL, CQL, AWR) to learn a policy $\pi(a_t|s_t)$ that mimics the expert's tool-use strategies. The key components of this environment are:

\begin{itemize}
    \item \textbf{State Space ($\mathcal{S}$):} The state $s_t$ at step $t$ is a comprehensive representation of the problem-solving history. It is a tuple $s_t = (I, Q, H_t)$, where $I$ is the input image, $Q$ is the initial user query, and $H_t = \{(a_0, o_0), \dots, (a_{t-1}, o_{t-1})\}$ is the history of previous action-observation pairs. The observation $o_i$ is the output returned by the tool invoked in action $a_i$.
    \item \textbf{Action Space ($\mathcal{A}$):} The action space is discrete and contains all available operations the agent can perform. It is the union of the tool set and a special `[Answer]` action: $\mathcal{A} = \{\text{tool}_1(\dots), \dots, \text{tool}_N(\dots)\} \cup \{[Answer](...)\}$. Each tool may take specific arguments (e.g., object IDs, coordinates), which the agent must generate.
    \item \textbf{Process Reward Function ($\mathcal{R}$):} Following process supervision principles \citep{lightman2023letsverifystepstep}, we implement dense, step-level rewards rather than sparse outcome-based rewards. We propose a multi-faceted reward function that combines multiple quality dimensions:
    $$ r_t = \alpha \cdot r_{\text{verifier}}(s_t, a_t) + \beta \cdot r_{\text{efficiency}}(a_t) + \gamma \cdot r_{\text{necessity}}(s_t, a_t) $$
    where $r_{\text{verifier}} \in [0, 10]$ is our automated quality assessment, $r_{\text{efficiency}}$ penalizes redundant tool calls, and $r_{\text{necessity}}$ rewards actions that meaningfully advance spatial understanding. This full function provides a rich learning signal. The empirical results in this paper validate the primary component of this signal ($r_{\text{verifier}}$), which serves as a strong proxy for overall reasoning quality.
    \item \textbf{Transition Dynamics ($\mathcal{P}$):} Since we are in an offline setting, the transition dynamics are deterministic and defined by the dataset. The next state $s_{t+1}$ is determined by the current state $s_t$ and the action $a_t$ taken in the recorded trajectory, where the tool output $o_t$ updates the history to $H_{t+1}$.
\end{itemize}

This format enables both offline RL and supervised fine-tuning applications (see Appendix~\ref{app:sft_details}).

\subsection{Core Components}
SpatialTraceGen's architecture consists of two primary components: a central reasoning agent and a modular suite of vision tools.

\subsubsection{Single Hop Generator}
The generator is a VLM agent that decomposes the user query $Q$ into a sequence of actions $\{a_0, a_1, \dots, a_T\}$. At each step $t$, it uses the current state $s_t$ to select the most appropriate tool from the tool suite to gather evidence, or outputs the final answer if it has sufficient information. The agent's behavior is guided by a system prompt that encourages strategic and efficient tool use (details in Appendix~\ref{app:prompts}).

\subsubsection{User-Registered Tool Suite}
The framework uses a "plug-and-play" architecture allowing easy integration of specialized vision models. This modularity also provides a key lever for efficiency, allowing a user to swap in faster, less computationally expensive tools during data generation to balance trace quality with generation cost. For our experiments, we use state-of-the-art models for three critical spatial capabilities: segmentation to identify and isolate objects \citep{ravi2024sam2}, depth estimation to infer distance and 3D relationships \citep{yang2024depth_anything_v2}, and 3D scene reconstruction to generate novel viewpoints \citep{xiang2024structured}. Visualizations are in Figure~\ref{fig:spatialtrace-tool-visuals}, with detailed descriptions in Appendix~\ref{app:tool_details}.

\subsubsection{Verifier}
A key innovation in our framework is the use of a Verifier LLM (GPT-4o \citep{openai2024gpt4o}) for automated quality control. After each action $a_t$, the Verifier assesses its quality and provides the primary reward signal, $r_{\text{verifier}}$, for our process reward function. The generation process uses this feedback, only accepting steps that meet a quality threshold $\tau$ or have undergone a maximum of $\alpha$ regeneration attempts. This filters out flawed reasoning steps, ensuring a high-fidelity final dataset. The full Verifier prompt and rubric are in Appendix~\ref{app:verifier_details}.

\begin{figure}[t]
  \centering
  \includegraphics[width=\linewidth]{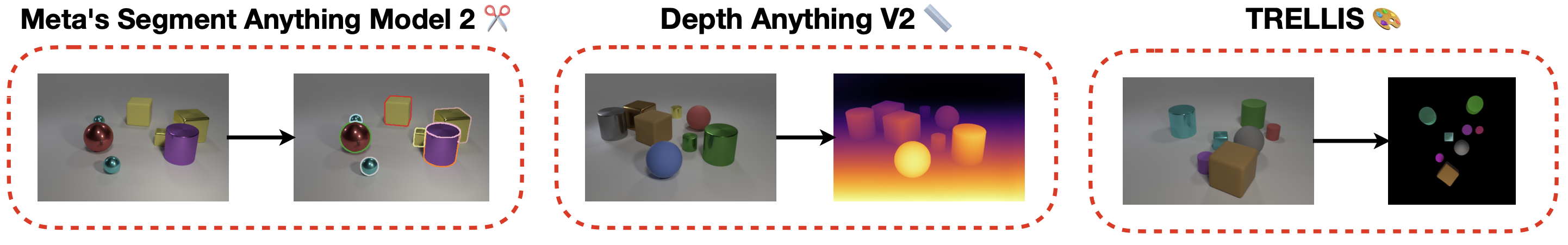}
  \vspace{-1.5em}
  \caption{\textbf{Tool visualizations using CLEVR-Humans sample images.} Our framework integrates diverse vision tools to provide rich, multi-modal information for spatial reasoning.}
  \label{fig:spatialtrace-tool-visuals}
  \vspace{-1em}
\end{figure}


\section{Experiments and Results}

\subsection{Experimental Setup}

SpatialTraceGen outputs structured JSON traces containing conversation history, verification details, tool outputs, and metadata for immediate SFT and offline RL compatibility (schema in Appendix~\ref{app:trace_schema}). We validate our framework via an ablation study on the CLEVR-Humans dataset \citep{johnson2017inferring}, generating traces under three conditions with varying verifier quality thresholds ($\tau = 0, 4, 5$) to isolate the verifier's impact on trace quality, consistency, and final answer accuracy. Full experimental details are in Appendix~\ref{app:exp_details}.


\subsection{Verifier-Driven Improvements in Data Quality}
Our results, summarized in Table~\ref{tab:quality_stats} and Figure~\ref{fig:quality_score}, show that our automated verifier significantly enhances the quality and consistency of the generated data, creating a more potent and sample-efficient training signal for downstream models. The primary findings demonstrate the efficiency of our approach: increasing the verifier threshold from $\tau=0$ (no verification) to $\tau=5$ (strict verification) raises the average trace quality score by 17\%. This process also reduces the standard error by nearly 50\% (from 0.054 to 0.028), indicating a substantial increase in the consistency and reliability of the reasoning traces. By filtering out low-quality reasoning steps, the verifier ensures that the generated dataset contains a stronger, more uniform learning signal.

\begin{table}[h]
\centering
\small 
\setlength{\tabcolsep}{4pt}
\begin{tabular}{cccc}
\toprule
Type of Verification & Threshold ($\tau$) & Accuracy & Quality Score \\
\midrule
None & 0 & $\textbf{74\%}$ & $6.508 \pm 0.054$ \\
Basic & 4 & $\textbf{74\%}$ & $7.499 \pm 0.056$ \\
Strict & 5 & $\textbf{74\%}$ & $\textbf{7.651} \pm \textbf{0.028}$ \\
\bottomrule
\vspace{1mm}
\end{tabular}
\caption{\textbf{Quantitative results on CLEVR-Humans.} Increasing the verifier threshold $\tau$ improves the average reasoning quality score and reduces its variance, all while maintaining final answer accuracy. Results are over 100 images.}
\label{tab:quality_stats}
\end{table}

\vspace{-1.25em}

Notably, the final answer accuracy remains constant at 74\% across all conditions. We hypothesize this is a ceiling effect from the benchmark's simplicity, where different reasoning paths can achieve the same correct answer. This confirms our framework's primary benefit is improving the reasoning process itself, which creates higher-quality demonstrations that provide a much stronger signal for distilling complex, multi-tool reasoning strategies.

\section{Conclusion}

We introduced SpatialTraceGen, a framework that efficiently generates high-fidelity, multi-tool reasoning traces for offline RL by using an automated Verifier for process-level supervision. Our experiments show this verifier-guided process improves trace quality by over 17\% and reduces variance by over 40\%, creating a more robust signal for sample-efficient learning. While final-answer accuracy on the CLEVR-Humans benchmark remained unchanged, we hypothesize this is due to a ceiling effect from the benchmark's simplicity, highlighting that our framework improves the reasoning process itself. The primary limitation is the upfront computational cost of generation, which we frame as a deliberate trade-off for enabling more efficient downstream training of smaller models. Future work will focus on using this dataset to fine-tune these smaller models and empirically validating their performance gains on more challenging spatial reasoning benchmarks, which may better reveal the benefits of high-quality reasoning.

\medskip

\newpage
{
\small
\bibliographystyle{plainnat}
\bibliography{references}

\begin{thebibliography}{16}
\providecommand{\natexlab}[1]{#1}
\providecommand{\url}[1]{\texttt{#1}}
\expandafter\ifx\csname urlstyle\endcsname\relax
  \providecommand{\doi}[1]{doi: #1}\else
  \providecommand{\doi}{doi: \begingroup \urlstyle{rm}\Url}\fi

\bibitem[Chen et~al.(2025)Chen, Zhu, Zhou, Zhang, Gao, Niebles, Geva, He, Wu,
  and Li]{chen2025spatialreasoninghardvlms}
Shiqi Chen, Tongyao Zhu, Ruochen Zhou, Jinghan Zhang, Siyang Gao, Juan~Carlos
  Niebles, Mor Geva, Junxian He, Jiajun Wu, and Manling Li.
\newblock Why is spatial reasoning hard for vlms? an attention mechanism
  perspective on focus areas, 2025.
\newblock URL \url{https://arxiv.org/abs/2503.01773}.

\bibitem[Feng et~al.(2025)Feng, Huang, Qu, Zhang, Qin, Zhong, Jiang, Chi, and
  Zhong]{feng2025retool}
Jiazhan Feng, Shijue Huang, Xingwei Qu, Ge~Zhang, Yujia Qin, Baoquan Zhong,
  Chengquan Jiang, Jinxin Chi, and Wanjun Zhong.
\newblock {ReTool}: Reinforcement learning for strategic tool use in llms,
  2025.
\newblock URL \url{https://arxiv.org/abs/2504.11536}.

\bibitem[Goldie et~al.(2025)Goldie, Mirhoseini, Zhou, Cai, and
  Manning]{goldie2025synthetic}
Anna Goldie, Azalia Mirhoseini, Hao Zhou, Irene Cai, and Christopher~D.
  Manning.
\newblock {Synthetic Data Generation \& Multi-Step RL for Reasoning \& Tool
  Use}, 2025.
\newblock URL \url{https://arxiv.org/abs/2504.04736}.

\bibitem[Johnson et~al.(2017)Johnson, Hariharan, van~der Maaten, Hoffman,
  Fei-Fei, Zitnick, and Girshick]{johnson2017inferring}
Justin Johnson, Bharath Hariharan, Laurens van~der Maaten, Judy Hoffman,
  Li~Fei-Fei, C.~Lawrence Zitnick, and Ross Girshick.
\newblock Inferring and executing programs for visual reasoning.
\newblock In \emph{Proceedings of the IEEE International Conference on Computer
  Vision (ICCV)}, pages 3008--3017, 2017.
\newblock \doi{10.1109/ICCV.2017.325}.

\bibitem[Lee et~al.(2025)Lee, Je, Park, Uy, Guibas, and
  Sung]{lee2025perspectiveaware}
Phillip~Y. Lee, Jihyeon Je, Chanho Park, Mikaela~Angelina Uy, Leonidas Guibas,
  and Minhyuk Sung.
\newblock {Perspective-Aware Reasoning in Vision-Language Models via Mental
  Imagery Simulation}, 2025.
\newblock URL \url{https://arxiv.org/abs/2504.17207}.

\bibitem[Lightman et~al.(2023)Lightman, Kosaraju, Burda, Edwards, Baker, Lee,
  Leike, Schulman, Sutskever, and Cobbe]{lightman2023letsverifystepstep}
Hunter Lightman, Vineet Kosaraju, Yura Burda, Harri Edwards, Bowen Baker, Teddy
  Lee, Jan Leike, John Schulman, Ilya Sutskever, and Karl Cobbe.
\newblock Let's verify step by step, 2023.
\newblock URL \url{https://arxiv.org/abs/2305.20050}.

\bibitem[Lu et~al.(2025)Lu, Chen, Liu, Thapa, Boen, and Zou]{lu2025octotools}
Pan Lu, Bowen Chen, Sheng Liu, Rahul Thapa, Joseph Boen, and James Zou.
\newblock {OctoTools}: An agentic framework with extensible tools for complex
  reasoning, 2025.
\newblock URL \url{https://arxiv.org/abs/2502.11271}.

\bibitem[Marsili et~al.(2025)Marsili, Agrawal, Yue, and
  Gkioxari]{marsili2025visualagentic}
Damiano Marsili, Rohun Agrawal, Yisong Yue, and Georgia Gkioxari.
\newblock {Visual Agentic AI for Spatial Reasoning with a Dynamic API}, 2025.
\newblock URL \url{https://arxiv.org/abs/2502.06787}.

\bibitem[Ogezi and Shi(2025)]{ogezi2025spare}
Michael Ogezi and Freda Shi.
\newblock {SpaRE}: Enhancing spatial reasoning in vision-language models with
  synthetic data, 2025.
\newblock URL \url{https://arxiv.org/abs/2504.20648}.

\bibitem[{OpenAI}(2024)]{openai2024gpt4o}
{OpenAI}.
\newblock {Hello GPT-4o}.
\newblock \\url{https://openai.com/index/hello-gpt-4o/}, May 2024.
\newblock Accessed: 2025-06-10.

\bibitem[Ravi et~al.(2024)Ravi, Gabeur, Hu, Hu, Ryali, Ma, Khedr, Rädle,
  Rolland, Gustafson, Mintun, Pan, Alwala, Carion, Wu, Girshick, Dollár, and
  Feichtenhofer]{ravi2024sam2}
Nikhila Ravi, Valentin Gabeur, Yuan-Ting Hu, Ronghang Hu, Chaitanya Ryali,
  Tengyu Ma, Haitham Khedr, Roman Rädle, Chloe Rolland, Laura Gustafson, Eric
  Mintun, Junting Pan, Kalyan~Vasudev Alwala, Nicolas Carion, Chao-Yuan Wu,
  Ross Girshick, Piotr Dollár, and Christoph Feichtenhofer.
\newblock {SAM 2: Segment Anything in Images and Videos}, 2024.
\newblock URL \url{https://arxiv.org/abs/2408.00714}.

\bibitem[Stogiannidis et~al.(2025)Stogiannidis, McDonagh, and
  Tsaftaris]{stogiannidis2025gap}
Ilias Stogiannidis, Steven McDonagh, and Sotirios~A. Tsaftaris.
\newblock {Mind the Gap}: Benchmarking spatial reasoning in vision-language
  models, 2025.
\newblock URL \url{https://arxiv.org/abs/2503.19707}.

\bibitem[Su and et~al.(2025)]{Su2025PixelReasoner}
Alex Su and et~al.
\newblock Pixel reasoner: Incentivizing pixel-space reasoning with
  curiosity-driven reinforcement learning.
\newblock \emph{{arXiv} preprint arXiv:2505.15966}, May 2025.
\newblock URL \url{https://arxiv.org/abs/2505.15966}.

\bibitem[Xiang et~al.(2024)Xiang, Lv, Xu, Deng, Wang, Zhang, Chen, Tong, and
  Yang]{xiang2024structured}
Jianfeng Xiang, Zelong Lv, Sicheng Xu, Yu~Deng, Ruicheng Wang, Bowen Zhang,
  Dong Chen, Xin Tong, and Jiaolong Yang.
\newblock Structured 3d latents for scalable and versatile 3d generation.
\newblock \emph{arXiv preprint arXiv:2412.01506}, 2024.

\bibitem[Yang et~al.(2025)Yang, Fu, Jia, Dong, Ma, Chen, Yang, and
  Wu]{yang2025spatialcognition}
Anran Yang, Cheng Fu, Qingren Jia, Weihua Dong, Mengyu Ma, Hao Chen, Fei Yang,
  and Hui Wu.
\newblock Evaluating and enhancing spatial cognition abilities of large
  language models.
\newblock \emph{International Journal of Geographical Information Science},
  2025.
\newblock \doi{10.1080/13658816.2025.2490701}.
\newblock URL \url{https://doi.org/10.1080/13658816.2025.2490701}.

\bibitem[Yang et~al.(2024)Yang, Kang, Huang, Zhao, Xu, Feng, and
  Zhao]{yang2024depth_anything_v2}
Lihe Yang, Bingyi Kang, Zilong Huang, Zhen Zhao, Xiaogang Xu, Jiashi Feng, and
  Hengshuang Zhao.
\newblock Depth anything v2.
\newblock \emph{arXiv preprint arXiv:2406.09414}, 2024.

\end{thebibliography}
}

\newpage

\appendix

\section{Technical Appendices and Supplementary Material}

\subsection{Supervised Fine-Tuning Formulation}
\label{app:sft_details}

In addition to its utility for offline reinforcement learning, the dataset generated by SpatialTraceGen is perfectly suited for Supervised Fine-Tuning (SFT). The goal of SFT in this context is to train a model to clone the expert's behavior through imitation learning.

We restructure the traces from our dataset into a collection of input-output pairs for standard language model training. For each step $t$ in a given trace $\gamma$, we create a training instance where the input is the state representation $s_t$ (formatted as a text prompt) and the target output is the expert's action $a_t$. The model is then trained to maximize the conditional probability of the expert's action given the state by minimizing the negative log-likelihood:
$$ \mathcal{L}_{SFT} = - \sum_{\gamma \in \mathcal{D}} \sum_{t=0}^{T} \log p(a_t | s_t) $$
This SFT approach provides an efficient way to initialize a policy with the expert's reasoning patterns. The high quality of our verifier-vetted traces is particularly crucial here, as noisy or suboptimal actions in the training data can significantly degrade the learned policy. The resulting SFT model can be deployed directly or used as a warm start for further refinement with offline RL algorithms.

\subsection{Tool Suite Implementation Details}
\label{app:tool_details}
\textbf{Segmentation.} To identify and isolate specific objects in a scene using bounding edges for precise object delineation, we utilize Meta's Segment Anything Model 2 (SAM 2) \citep{ravi2024sam2}. This work extends the original SAM with unified video and image segmentation capabilities, providing real-time visual segmentation with 6x faster performance and superior accuracy through streaming memory architecture for temporal consistency. 

\textbf{Depth Estimation.} To generate dense distance images that capture precise depth information across the entire scene, we employ Depth Anything V2 (DAv2) \citep{yang2024depth_anything_v2}, which produces significantly more robust and fine-grained depth predictions than its predecessor. It uses a three-stage approach using synthetic data, scaled teacher models, and large-scale pseudo-labeled real images, offering models ranging from 25M to 1.3B parameters for diverse computational requirements.

\textbf{3D Scene Reconstruction.} To generate comprehensive 3D spatial representations from single images for enhanced geometric understanding, we leverage TRELLIS \citep{xiang2024structured}, a unified 3D generation model that employs Structured LATent (SLAT) representations with rectified flow transformers. TRELLIS allows images to be rendered from any viewpoint including top-down perspectives for spatial analysis, which is the viewpoint we focus on in our tool call.

\subsection{System Prompts} \label{app:prompts}
The central agent of the framework is an lightweight Large Vision Language Model (e.g., Llama, Gemma), which acts as the "brain" of the operation. We choose GPT-4o \citep{openai2024gpt4o} as our Single Hop Generator. We encourage the Generator to balance diverse and strategic tool use. For example, we ensure that the Generator does not call the same tool repeatedly, but we also encourage it to gather the most relevant information needed to solve the problem. We also encourage increased information collection in order to increase answer confidence and certainty. 

Single Hop Generator System Prompt:
\begin{Verbatim}[breaklines=true,breakanywhere=true]
You are an expert AI in spatial reasoning. Your goal is to solve a user's question about an image by generating a step-by-step reasoning trace.

You have access to a suite of tools:
1. `trellis`: A bird's eye view tool. Call this to understand relative relationships between objects and identify objects. Returns a top-down view of the image. Note that the BOTTOM of the tool output image is the FRONT, and the TOP is the BACK. The LEFT and RIGHT are the same as normal.
2. `sam2`: A segmentation tool. Returns the image with each object is outlined with a colored borde. Call this to identify and outline objects in the image.
3. `dav2`: A depth estimation tool. Returns the image colorcoded to the depth of each part of the image. Call this to understand the relative distances of objects from the camera.

At each step, your response MUST be a single, valid JSON object with BOTH reasoning and an action. Do not add any explanatory text outside of the JSON structure.

Each response must include:
1. "reasoning": Your thought process for this step
2. "action": Either "tool_call" or "final_answer"
3. Additional required fields based on the action:

For tool calls:
{
  "reasoning": "Explain why you need to use this tool and what you expect to learn",
  "action": "tool_call",
  "tool_name": "trellis" or "sam2" or "dav2"
}

For final answers:
{
  "reasoning": "Explain your final reasoning based on all previous steps",
  "action": "final_answer",
  "text": "your_final_answer_here"
}

The possible answer choices are large, small, cube, cylinder, sphere, rubber, metal, gray, blue, brown, yellow, red, green, purple, cyan, yes, no, or a singular integer.
Note for final answer text, you MUST answer with ONE of the possible answer choices. 

Always provide clear reasoning that explains your thought process before taking the action.

Guidelines for effective spatial reasoning:
- Start by understanding what objects and spatial relationships the question asks about
- Use tools when you need to better understand the scene (segmentation, depth)
- Reason step by step, building up your understanding
- Be precise in your final answer - match the expected format (Yes/No, number, etc.)
- If you're uncertain, use tools to gather more information before concluding 
- MAXIMIZE the step by step thinking following each tool call output. Think CRITICALLY and CAREFULLY from multiple tool call sources.
- Call each tool once, rather than repeatedly calling the same tool. 
- After each tool call, please DESCRIBE the IMAGE thoroughly before providing an reasoning.
\end{Verbatim}

Verifier System Prompt:
\begin{Verbatim}[breaklines=true,breakanywhere=true]
# Verifier System Prompt for SpatialTraceGen

You are an expert verifier for spatial reasoning traces, ensuring Vision-Language Models learn optimal spatial cognition through high-quality training data.

## Core Mission

Evaluate each reasoning step to ensure it demonstrates **accurate, thorough spatial investigation** that prioritizes the best available information before drawing conclusions.

## Evaluation Criteria

### Information Quality (Priority #1)
- Does this step gather the most accurate, reliable information available?
- Are the chosen tools capable of providing the precision needed?
- Would more authoritative tools significantly improve reliability?

### Reasoning Excellence
- Is the logic sound and would experts agree?
- Does the step meaningfully advance spatial understanding?
- Are tool selections optimal for the stated sub-goal?

### Investigative Thoroughness
- Does this demonstrate comprehensive spatial exploration?
- Are multiple complementary tools leveraged effectively?
- Would additional tools provide valuable cross-validation?

## Philosophy: Quality Over Convenience

- **Accuracy First**: Always favor steps that ensure superior information quality
- **Multiple Tools Add Value**: Different tools reveal complementary spatial perspectives
- **Cross-Validation Builds Confidence**: Verify findings through diverse approaches
- **Methodical > Quick**: Better to investigate thoroughly than rush to conclusions
- **Be Generous**: Recognize that thorough investigation often requires multiple approaches and tool combinations

## Output Format

```json
{
  "necessity_analysis": "Whether this step meaningfully advances spatial understanding and contributes valuable information to the investigation",
  "correctness_analysis": "Assessment of reasoning soundness and appropriateness of tool selection for the spatial task",
  "efficiency_analysis": "Evaluation of whether this approach balances thoroughness with practical investigation methods",
  "alternative_approaches": "Other tools or methods that could complement this step or provide additional valuable perspectives",
  "critical_concerns": "Any significant issues with reasoning, tool usage, or potential for misleading conclusions",
  "rating": 7,
  "rating_justification": "Clear explanation for the rating, considering information quality and investigative value",
  "regeneration_needed": true,
  "suggested_improvement": "Specific guidance for enhancing the step's contribution to spatial understanding"
}
```

## Rating Scale (1-10)

- **1-3**: Significantly flawed reasoning, inappropriate tools, or misleading information
- **4-6**: Basic contribution but could benefit from better tools or more thorough investigation  
- **7-8**: Solid spatial reasoning with good information gathering and meaningful progress
- **9-10**: Exemplary demonstration of comprehensive, accurate spatial analysis

## Key Insight

The best spatial reasoning traces teach models to be **information maximalists** - systematically gathering high-quality data through thoughtful tool combinations. Value steps that demonstrate rigorous, evidence-based approaches to spatial problem-solving, even when they take a more exploratory path to build comprehensive understanding.
\end{Verbatim}

\label{app:verifier_details}
The \textit{Verifier} sends in-context feedback to the Single Hop Generator about the current step. In this work, we choose GPT-4o \citep{openai2024gpt4o} as our Verifier. The feedback is in terms of a score from 1-10, where 10 indicates an excellent step, and 1 indicates a poor or incorrect step. The Verifier has two hyperparameters: the minimum acceptance threshold $\tau$ and the maximium regeneration attempts (per step) $\alpha$. If the feedback score is at least as large as $\tau$, then the pipeline proceeds to the next step. Otherwise, it regenerates the current step, unless we have already used $\alpha$ regeneration attempts at this step in which case we also proceed.

\subsection{The Iterative Trace Generation Process}
\label{app:generation_process}
The generation of a single reasoning trace is illustrated in Figure 1 in the main paper. First, the Single Hop Generator receives the initial query and the visual context. It reasons about the logical sub-goals (e.g., "First, I need to locate the red car and the blue truck."), and selects the appropriate tool to achieve this. For example, to locate the car and truck, it might choose to use a segmentation tool (SAM 2). The Generator then prepares the input for the selected tool and invokes it. The raw output from the tool (e.g., segmentation bounding box) is processed into a usable format and sent back to to the Single Hop Generator. The generated step is then sent to the Verifier. The Verifier grades the step from 1-10, and this feedback is sent back to the Generator, which uses it to decide whether to proceed to the next step, or regenerate the current step. The results from previous steps are integrated into the cumulative understanding of the Single Hop Generator and are used to formulate the next sub-goal. This cycle continues until a final answer is arrived at. The collection of steps taken to reach this answer are aggregated into a final multihop trace, and added to the SpatialTraceGen dataset.

\subsection{Detailed Trace Structure (JSON Schema)}
\label{app:trace_schema}
The final output of our pipeline is a comprehensive structured trace in JSON format containing four key components:

\textbf{Conversation Trace:} The core \texttt{trace} field stores the complete multi-turn conversation as message arrays with \texttt{role} and \texttt{content}. Assistant messages contain structured JSON responses with \texttt{reasoning}, \texttt{action} (\texttt{tool\_call} or \texttt{final\_answer}), and action-specific parameters. User messages include textual tool outputs and encoded images from vision tools.

\textbf{Verification History:} Each reasoning step undergoes evaluation stored in \texttt{verification\_history}, recording \texttt{step\_index}, \texttt{attempt\_number}, \texttt{timestamp}, and comprehensive \texttt{result} analysis including necessity assessment, correctness evaluation, efficiency analysis, numerical \texttt{rating}, and \texttt{rating\_justification}. The system tracks whether steps \texttt{passed\_threshold} and if \texttt{regeneration\_triggered}.

\textbf{Tool Image Tracking:} The \texttt{tool\_images} array logs all vision tool outputs with \texttt{step\_index}, \texttt{attempt}, \texttt{tool\_name}, local file paths, invocation \texttt{reasoning}, and timestamps, enabling complete reproducibility of the multi-modal reasoning process.

\textbf{Metadata:} Additional fields capture the original \texttt{question}, \texttt{expected\_answer}, \texttt{difficulty} level, \texttt{average\_rating} across verification steps, and \texttt{generation\_timestamp}. This structure enables comprehensive analysis of reasoning quality, tool usage patterns, and verification effectiveness across different question types and difficulty levels.

\subsection{Experimental Setup Details}
\label{app:exp_details}

\begin{wrapfigure}{l}{0.4\textwidth}
   \vspace{-15pt}
   \centering
   \includegraphics[width=\linewidth]{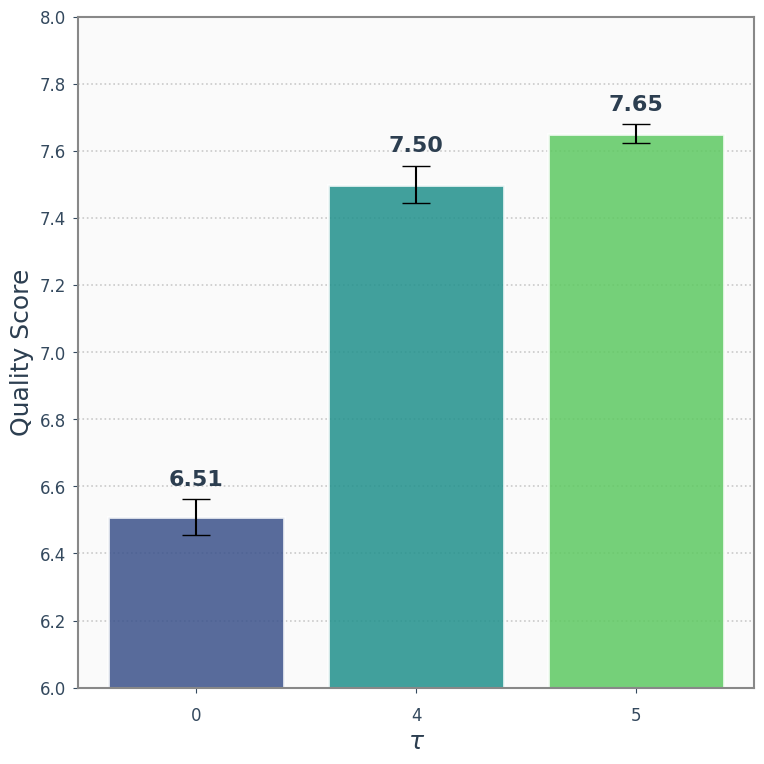}
   \captionsetup{width=0.25\textwidth}
   \caption{\textbf{Impact of verification threshold $\tau$ on reasoning quality scores}}
   \label{fig:quality_score}
   \vspace{-5pt}
\end{wrapfigure}

\textbf{Dataset and Tools.} We utilize a curated subset of 100 images from the the CLEVR-Humans dataset \citep{johnson2017inferring} containing human-annotated spatial reasoning questions of varying difficulty levels. Each sample includes a synthetic 3D scene image, a natural language question requiring multi-hop spatial reasoning, and a ground truth answer. Our framework employs GPT-4o as the Single Hop Generator, augmented with three specialized vision tools: Segment Anything Model 2 (SAM 2) \citep{ravi2024sam2} for object segmentation, Depth Anything V2 (DAv2) \citep{yang2024depth_anything_v2} for monocular depth estimation, and TRELLIS \citep{xiang2024structured} for 3D scene reconstruction and novel view synthesis. For SAM 2, we use the bounding edges of objects, and for TRELLIS, we use the bird's eye view (top-down).

\textbf{Verification System Design.} We use an independent LLM evaluator as our Verifier that assesses each reasoning step using a 10-point scale with clearly defined criteria: ratings 1-3 indicate significantly flawed reasoning with inappropriate tool selection; ratings 4-6 represent basic but improvable contributions; ratings 7-8 demonstrate solid spatial reasoning with effective information gathering; and ratings 9-10 exemplify comprehensive, accurate spatial analysis. When a step receives a rating below the specified threshold, the system triggers regeneration with a maximum of $\alpha$ times per step to prevent excessive computational overhead. We typically choose $\alpha = 2$ for our experiments.

\textbf{Experimental Conditions.} We design three experimental conditions to systematically evaluate the impact of verification-guided quality control:

\begin{enumerate}
\item \textbf{Without Verification} ($\tau = 0$): Serves as our baseline condition where traces are generated without any quality assessment or regeneration mechanism. This condition captures the raw performance of our multi-tool orchestration framework.

\item \textbf{With Basic Verification} ($\tau = 4$): Implements our verification system with a lenient quality threshold that accepts traces demonstrating "basic contribution but could benefit from better tools or more thorough investigation." This condition allows us to measure the impact of minimal quality filtering.

\item \textbf{With Strict Verification} ($\tau = 5$): Employs a more stringent quality threshold that requires traces to exceed basic-level reasoning. This condition targets traces that demonstrate meaningful spatial analysis capabilities.
\end{enumerate}

\textbf{Implementation Details.} We deliberately chose relatively moderate quality thresholds ($\tau = 4, 5$) after preliminary experiments revealed that excessively high thresholds ($\tau \geq 6$) resulted in traces trapped in continuous regeneration loops. Each verification threshold processes identical input samples to ensure fair comparison, with all generated traces, verification histories, and intermediate tool outputs systematically logged for comprehensive analysis.

\subsection{Additional Dataset Analysis and Examples}
\label{app:trace_examples}

\subsubsection{Trace Distribution}
\begin{figure}[h]
\centering
\includegraphics[width=0.8\linewidth]{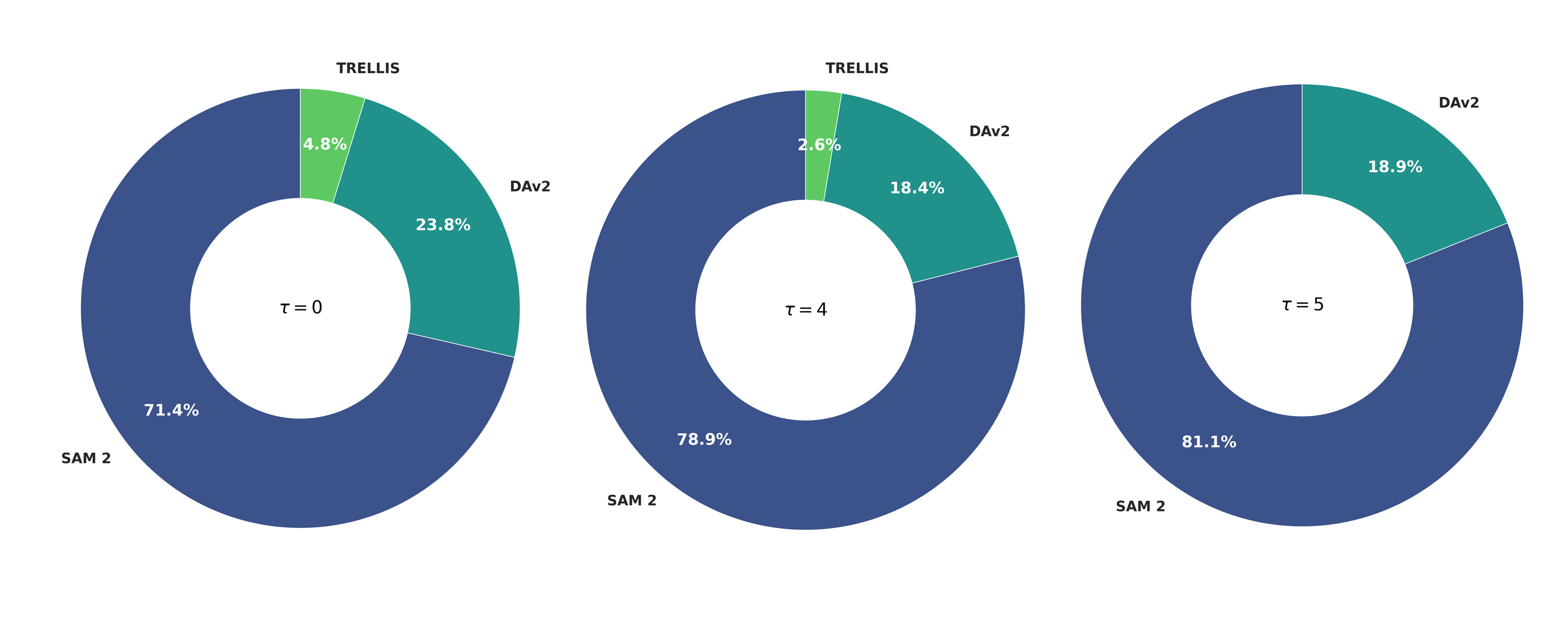}
\caption{\textbf{Distribution of tool calls over different minimum acceptance thresholds ($\tau$)}}
\label{fig:tool_distributions}
\end{figure}

We note that our distribution of tool calls is heavily skewed towards SAM 2, and away from TRELLIS (Figure~\ref{fig:tool_distributions}). We also observe that the use of the Verifier decreases the emphasis of TRELLIS. We attribute this to the CLEVR dataset, which consists of several objects that must be classified in order to answer the spatial questions. Thus, this tool is the most useful in this problem setting. In another problem setting, TRELLIS or DAv2 may be more useful (e.g., if we have to estimate exact distances between objects).

\subsubsection{Qualitative Trace Examples}
\begin{figure}[h]
\centering
\includegraphics[width=\linewidth]{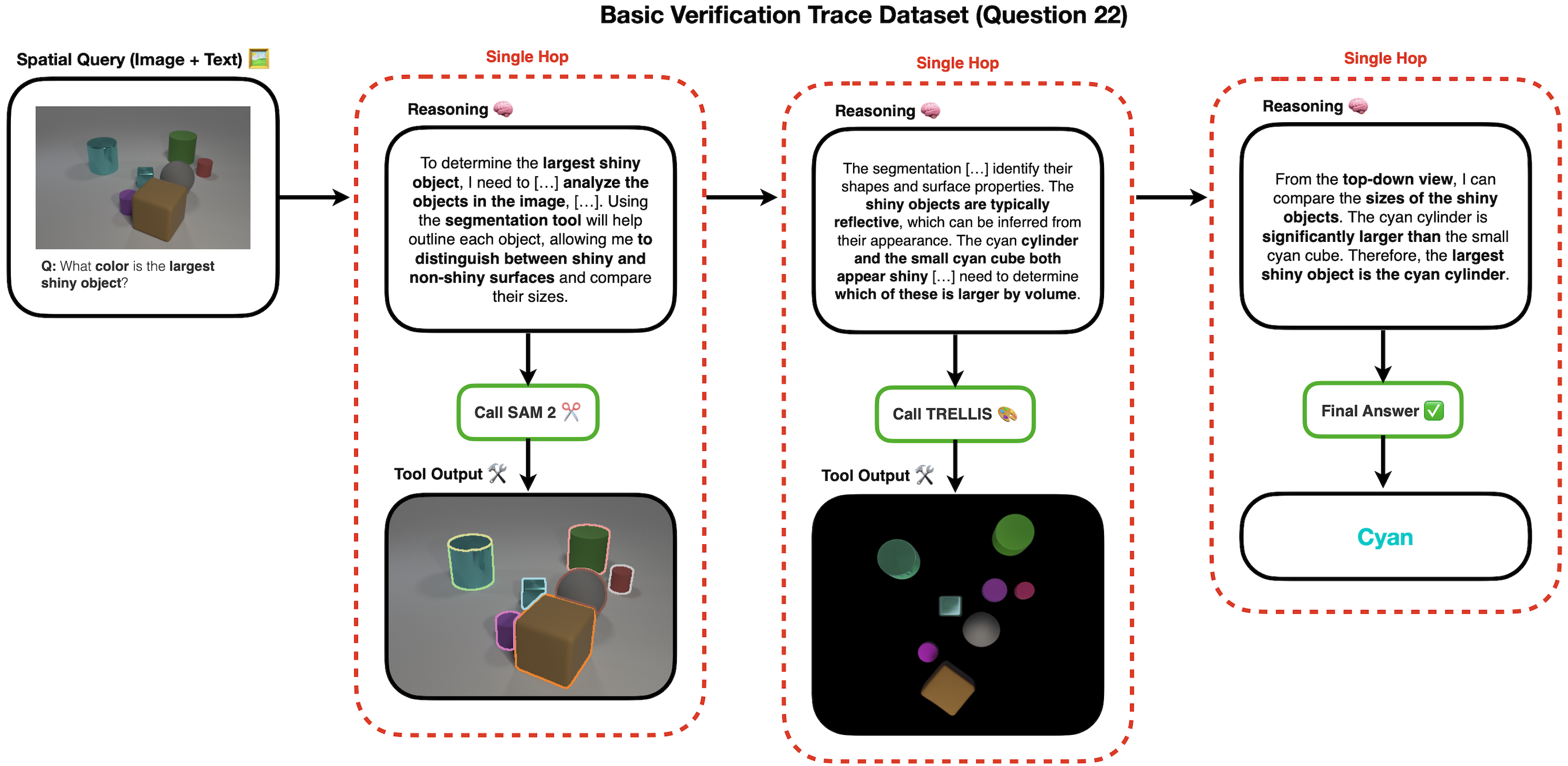}
\caption{\textbf{Example trace for $\tau = 4.0$.} Reasoning performed on question 22 (basic verification).}
\label{fig:example_trace_basic}
\end{figure}

\begin{figure}[h]
\centering
\includegraphics[width=\linewidth]{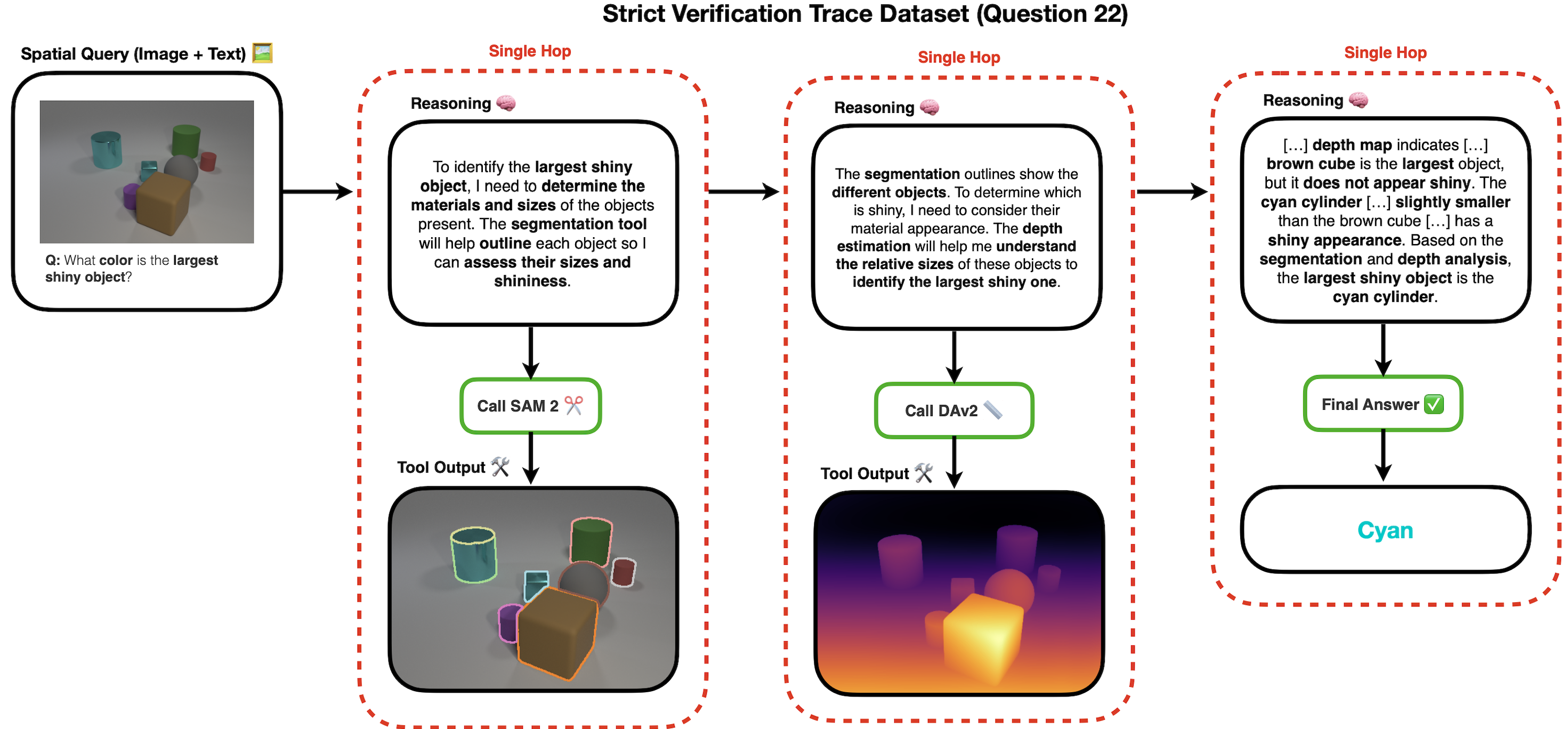}
\caption{\textbf{Example trace for $\tau = 5.0$.} Reasoning performed on question 22 (strict verification).}
\label{fig:example_trace_strict}
\end{figure}

Figures~\ref{fig:example_trace_basic} and \ref{fig:example_trace_strict} illustrate how verification pressure drives strategic tool diversification. Both traces attempt the question ``What color is the largest shiny object?'' and correctly identify the cyan cylinder without access to ground truth during generation. While both employ SAM 2 for initial object segmentation, they diverge in addressing depth bias affecting apparent object sizes. The basic verification trace (Figure~\ref{fig:example_trace_basic}) leverages TRELLIS's top-down view to eliminate perspective distortion entirely, reasoning that ``from the top-down view, I can compare the sizes of the shiny objects'' without depth interference. Conversely, the strict verification trace (Figure~\ref{fig:example_trace_strict}) employs DAv2's depth estimation to explicitly measure and correct for distance effects, noting that the ``depth map indicates [that the] brown cube is the largest object, but it does not appear shiny.'' Thus, our framework recognizes that depth bias can be addressed through either perspective transformation or explicit depth measurement.

\section{Limitations}

Our pipeline inherently requires many API calls to the Single Hop Generator and the Verifier, which can grow computationally expensive with multiple steps or a stricter acceptance threshold with higher maximum regeneration attempts. Furthermore, the context length for each call grows significantly with the number of hops in a trace, as a single hop generation is conditioned on all previous hops. Thus, another bottleneck to the framework is the context length of the LLM agent. To mitigate this in our current implementation, we set the API call's detail to be "medium" for tool output. For better performance, a model with longer context length should be used, allowing for higher image detail. This can impose limitations on our choices of the Single Hop Generators and the Verifiers to cost-effective, smaller models. 


\end{document}